\documentclass[runningheads]{llncs}
\usepackage[utf8]{inputenc}
\usepackage{booktabs}
\usepackage{graphicx}
\usepackage[scaled=0.9]{couriers}
\usepackage{xcolor}
\usepackage{ulem}
\usepackage{microtype}
\usepackage{multirow}
\usepackage{hyperref}

\title{Distributional Analysis of \\ Polysemous Function Words}
\author{Sebastian Padó\inst{1}
\and Daniel Hole\inst{2}
}

\institute{Natural Language Processing (IMS), University of Stuttgart, Germany \and Linguistics (IL), University of Stuttgart \\
\email{\{sebastian.pado@ims, daniel.hole@ling\}.uni-stuttgart.de}}
\date{}

\begin{document}

\maketitle

\pagestyle{empty} 

\begin{abstract}
  In this paper, we are concerned with the phenomenon of function word
  polysemy. We adopt the framework of distributional semantics, which
  characterizes word meaning by observing occurrence contexts in large
  corpora and which is in principle well situated to model polysemy.
  Nevertheless, function words were traditionally considered as impossible
  to analyze distributionally due to their highly flexible usage
  patterns.

  We establish that \textit{contextualized word embeddings},
  the most recent generation of distributional methods, offer hope in
  this regard. Using the German reflexive pronoun \textit{sich} as an
  example, we find that contextualized word embeddings capture
  theoretically motivated word senses for \textit{sich} to the extent
  to which these senses are mirrored systematically in linguistic
  usage.
    
\end{abstract}

\section{Introduction} 

Theoretical linguists observe with envy the way in which distributional semantics in computational linguistics renders research viable whose foundations were postulated by clear-sighted structuralists  \cite{Firth1957,harris1954distributional}. Their interest diminishes upon seeing that computational linguistics has dealt mainly with parts of speech dominated by content words (nouns, verbs, adjectives), whereas theoretical linguists firmly believe that function words and morphosyntax define the interesting backbone of natural language. In this respect, the focus of computational linguistics has broadened only in recent years.

This paper brings together the advanced computational tools of
distributional semantics with the interest of formal linguistics in
function words and in particular their disambiguation. We consider a
multiply polysemous function word, the German reflexive pronoun
\textit{sich}, and investigate in which ways natural subclasses of
this word which are known from the theoretical and typological
literature map onto recent models from distributional semantics.  Due
to the differences between lexical and functional polysemy, our
results are different from those of distributional studies of
systematic polysemy in content words such as \cite{J12-3005}.

We submit that our results open a window onto patterns of polysemy
that may, in the long run, turn out at least as interesting and
relevant to the computational study of natural languages as content
words. What we find in our pilot is that some traditional subclasses
of \textit{sich} not only map neatly onto clusters produced by
distributional methods, but that others which are predicted by theory
to belong to constructional metaclasses with a wider distribution
pervade the whole clustering space. What is more, the distribution of
causative-transitive vis-à-vis anticausative verb types and of other
verb classes partly reproduces the semantic map of the middle domain
on a typological database \cite{Kemmer-1993}. We take these results to
be a promising starting point for more in-depth studies of function
morphemes in distributional semantics.

\section{Background: Distributional  Analysis}

Today, distributional analysis is the dominant paradigm for semantic
analysis in computational linguistics. Building on the distributional
hypothesis, \textit{``you shall know a word by the company it keeps"}
\cite{Firth1957}, it typically represents words as high-dimensional
vectors which summarize the words' occurrence contexts (see
\cite{Turney2010} for an introduction and overview). Traditionally,
these vectors were obtained by counting: each dimension corresponded
to one particular linguistic context (often, another word), and the
value in the vector for this dimension was the co-occurrence frequency
of the two words, or some function thereof.

This procedure was increasingly replaced by neural network-based
methods, where the co-occurrence frequencies are not directly used as
vectors. Instead, they form the ``output'' that the neural network is
supposed to predict, and the vectors are given by the internal
parameters of the neural network, now often called `word embeddings'
\cite{baroni-dinu-kruszewski:2014:P14-1}. Crucially, traditional
fundamental intuitions about distributional semantics mostly carry
over to the new paradigm. In fact, some widely used types of word
embeddings are mathematically equivalent to count vectors to which
dimensionality reduction has been applied \cite{levy2014neural}.

%
%
%


At the same time, the move to neural network-based vector learning has
opened the door for innovative network architectures. Prominent among
these are the recently introduced \textit{contextualized embeddings}.
These models concurrently learn (a) general vectors for word types
(lemmas) and (b) specialized vectors for word tokens (instances) in
their local context. In this manner, they overcome the traditional
limitation of distributional semantics, which generally used to
aggregate the contexts of all instances, and thus all senses, into one
vector. The most successful model architecture to create
contextualized embeddings are so-called \textit{transformers}
\cite{vaswani2017attention}, a class of models which lets each context
word directly influence the representation of the target word, and
automatically learns to weigh these contributions using a mechanism
called \textit{self-attention}.  In this process, which is carried out
several times, transformers uncover (some degree of) implicit
linguistic structure such as predicate-argument relations,
coreference, or phrase structure \cite{jawahar-etal-2019-bert}.

As introduced above, the focus of this paper is the polysemy of
function words such as \textit{sich}.  Traditionally, distributional
analysis has concentrated mostly on \textit{content} words (common
nouns, verbs and adjectives), following the intuition that these word
classes refer to categories whose properties and relational structure
can be learned from distributional analysis
\cite{cimiano2005learning}. Exceptions notably include distributional
studies of compositionality, which have modeled the semantic effects
of quantifiers \cite{bernardi-etal-2013-relatedness} and determiners
\cite{10.5555/2380816.2380822} on sentence-level entailment.

Crucially, these studies do not consider polysemous function
words. Indeed, the context of function words is typically so general
that traditional methods of distributional analysis tended to fail in
this domain, since any reflection of the function word meaning was
likely to be masked by the topic of the surrounding linguistic
material. Consequently, the only (partially functional) word category
that has received more than cursory attention in distributional
semantics with regard to senses and disambiguation are prepositions
\cite{bannard2003distributional,schneider-preps-acl18}. Our study
takes benefit of the development that the contextualized embeddings
created by transformers take a major step towards alleviating the
generality problem: Even if the representation of the word type
\textit{sich} is still too general to be useful, the embeddings for
each instance of \textit{sich}, arising from the combination of word
type meaning and context, is informative enough for analysis.


\section{Phenomenon: The German reflexive pronoun \textit{sich}}

\begin{table}[t!]
\centering
\setlength{\tabcolsep}{3pt}
\begin{tabular}{p{9cm}ccccc}
\toprule
    Class / Example & \rotatebox{90}{\begin{minipage}{1.6cm} predictable \end{minipage}} & \rotatebox{90}{\begin{minipage}{1.6cm} agentive \end{minipage}}  & \rotatebox{90}{\begin{minipage}{1.6cm} stressable \end{minipage}} & \rotatebox{90}{\begin{minipage}{1.6cm} +\textit{lassen} \end{minipage}}  &\rotatebox{90}{\begin{minipage}{1.6cm}disposition \end{minipage}}  \\ 
    \midrule
\textsc{1. Inherent reflexives}:
\textit{Paul schämte sich}/ \\ `Paul felt ashamed'
& +
& $\pm$
& -
& -
& $\pm$ \\
\textsc{2. Anti-causatives}: 
\textit{Die Erde dreht sich}/ \\ `The earth revolves'
& +
& -
& -
& - 
& $\pm$ \\
\textsc{3. Change in posture}:
\textit{Paul setzte sich hin}/ \\ `Paul sat down'
& +
& $\pm$
& -
& -
& - \\
\textsc{4. Typically self-directed}:
\textit{Paul kämmte sich}/ \\ `Paul combed his hair'
& -
& +
& -
& -
& - \\
\textsc{5. Typically other-directed}:
\textit{Paul erschoss sich}/ \\ `Paul shot himself'
& -
& +
& + 
& -
& - \\
\textsc{6. Dispositional middle}:
\textit{Die Dose lässt sich leicht öffnen}/ \\ `The can opens easily'
& +
& -
& -
& +
& + \\
\textsc{7. Episodic middle}:
\textit{Paul lässt sich beraten}/ \\ `Paul takes advice'
& +
& +
& -
& +
& - \\
\textsc{8. Reciprocals}:
\textit{Die Geraden schneiden sich im Unendlichen}/ \\ `The lines intersect in the infinite'
& -
& $\pm$
& $\pm$
& -
& $\pm$ \\
\bottomrule
\end{tabular}
\caption{Salient classes of \textit{sich}, inspired by Kemmer (1991), plus feature representation ($\pm$ indicates the possibility of both positive and negative cases depending on context)}
\label{tab:kemmer}
\end{table}

The reflexive pronoun in German is a notorious case of polysemy
because prototypical instances such as \textit{sich loben} `praise
oneself' are by far outnumbered by other uses. These other uses cover
large portions of what has come to be known as the `middle domain' in
linguistic typology \cite{Kemmer-1993}. The classification in
Table~\ref{tab:kemmer} provides our simplified overview of this domain
in German including examples.

Class 1 is a metaclass, as it assembles historically fossilized
combinations of verbs with reflexive pronouns (\textit{sich benehmen}
`to behave oneself'). These verbs invariably occur with reflexive
pronouns. This class includes fossilized combinations of \textit{sich}
with prepositions, such as Kant's \textit{Ding an sich} `thing in
itself'. The anti-causatives of Class 2 derive non-agentive
intransitive uses of transitive verbs (\textit{sich drehen} `to
turn'), potentially expressing a disposition. Class 3 comprises
constructions denoting changes in body posture with obligatory
\textit{sich}, such as \textit{sich setzen} `to sit down'.  Class 4
consists of agentive predicates such as predicates of grooming
(\textit{sich kämmen} `to comb one's hair') or predicates of
assessment (\textit{sich in der Lage sehen} `to feel equal to doing
sth.') which are typically, but not exclusively, used with
\textit{sich}. The `prototypical' \textit{sich} instances
(\textit{sich erschie\ss{}en} `to shoot oneself'), where \textit{sich}
is used to express the identity of subject and another arugment, are
concentrated in Class 5. Another diagnostic to distinguish classes 4
and 5 is that \textit{sich} is typically unstressed in Class 4,
whereas the reflexives of Class 5 may be stressed. The dispositional
middles of Class 6 form a construction that encodes a disposition of
the subject referent (\textit{sich leicht öffnen lassen} `to open
easily'). Class 7 is similar, but an episodic event is referred to
instead of the stative property of Class 6 (\textit{sich beraten
  lassen} `to get advice'). Class 8, finally, encompasses uses of
\textit{sich} that could be replaced by \textit{einander} `each other,
one another' and are, hence, reciprocals (\textit{sich kennen} `to
know each other').

One caveat is in order here. The classes are not completely mutually
exclusive. If, for instance, \textit{sich legen} `lay down' is used as
in \textit{...der sich wie eine weiße Schimmelschicht auf die Kleidung
  legt...} `...which covers the clothes like a white layer of mold',
either Class 2 or Class 3 (with a non-literal use) could host this
example. We avoided multiple classifications and allotted examples of
this kind on a `best fit' basis (Class 2 for the example given).

As the right hand hand side of the table shows, these eight senses can be distinguished in terms of five properties:
\begin{itemize}
\item Is \textit{sich} \textbf{predictable} in this context?
  Predictability is meant to describe the property that the reflexive
  pronoun in the relevant classes cannot be replaced by another 3rd
  person pronoun (\textit{$^{*}$ Paul schämte ihn}).
\item Is the event \textbf{agentive}?
    \item Is \textit{sich} stressable in this context?
    \item Does the construction involve \textit{lassen}?
    \item Does the construction describe a \textbf{disposition}?
\end{itemize}
In the table, the value $\pm$ indicates neutrality (both positive and
negative values exist, depending on context). In our experience, these
features can provide valuable criteria for choosing the right category
in manual annotation.

\section{Data and Annotation}
\label{sec:data-annotation}

As basis of our study, we use the 700M token SdeWAC web corpus
\cite{faas-13}. We selected the first 335 out of more than 5.5 million
instances of \textit{sich} for manual annotation with the eight
classes as defined above. The annotation was carried out by the two
authors individually. We computed Cohen's kappa as a measure of
inter-annotator agreement and obtained a value of 0.73, which
indicates substantial agreement \cite{kappaInterpretation}, despite
the possible non-exclusivity of the classes.

The confusion matrix is shown in Table~\ref{tab:confusion}. The
largest classes, according to both annotators, are Class 1, 2, and
4. There is essentially perfect agreement on the reciprocals and the
middles and some disagreement on Classes 4 and 5 (typically self-
vs. other-directed), but most diagreements involve Classes 1 through 3
-- specifically Class 1 vs. Class 2 (31 cases -- more than half of all
disagreements), and Class 1 vs. Class 3 (8 cases).

Some of the disagreements were oversights by one of the two
annotators.  However, there were also cases of systematic differences
in judgments. For instance, the Class 1 vs. Class 2 disagreements
often concern instances where the main criterion for Class 2
(intransitive use of transitive verb) is debatable:
\begin{quote}
  Jedes Jahr wieder \uline{stauen} \textit{sich} zur Urlaubszeit die
  Blechlawinen auf den Autobahnen [...]

  'Every year again, avalanches of metal \uline{back up} \textit{\O} on the
  motorways during holiday time, [...]'
\end{quote}
If one is willing to accept this as a reflexive analogue to transitive
uses like \textit{Blockaden stauen den Verkehr} 'blockades back up
traffic', this is a case of Class 2, otherwise Class 1.

As for Class 1 vs. Class 3, a recurring problem is to delineate the
verbs of change of posture (Class 3) -- in particular with regard to
nonliteral uses, which are frequent for motion verbs.  For example,
\begin{quote}
  Die Revision \uline{wendet} \textit{sich} nur gegen die Ansicht des Berufungsgerichts [...] .

  'The revision only \uline{turns against / opposes} \textit{\O} the view of the appellate court [...]'
\end{quote}

\begin{table}[t!]
  \centering
  \setlength{\tabcolsep}{3pt}
    \begin{tabular}{rcrrrrrrrrr}
    \toprule
 &   \multicolumn{8}{c}{Annotator 1} \\
& Class  & 1 &  2 &   3&   4&   5&   6&   7&   8 \\ \midrule
&1  &143&   6&   6&   1&   0&   0&   0&   0 \\
&2  & 25&  60&   0&   2&   0&   0&   0&   0 \\
\multirow{5}{0.5cm}{\rotatebox{90}{Annotator 2}}  &3  &  2&   1&  11&   0&   0&   0&   0&   0 \\
&4  &  6&   0&   1&  28&   4&   0&   0&   0 \\
&5  &  2&   0&   1&   3&  18&   0&   0&   0 \\
&6  &  0&   0&   0&   0&   0&   3&   0&   0 \\
&7  &  0&   0&   0&   0&   0&   0&   8&   0 \\
&8  &  1&   0&   0&   0&   0&   0&   0&   3 \\
    \bottomrule
    \end{tabular}
    \caption{Confusion matrix for \textit{sich} categories by two
      annotators}
    \label{tab:confusion}
\end{table}

We resolved these disagreements via joint adjudication.  The resulting
frequency distribution over classes is shown in
Table~\ref{tab:class-freqs}. The final labeled dataset is available,
together with the Jupyter notebook documenting the subsequent
analysis, from
\url{https://www.ims.uni-stuttgart.de/forschung/ressourcen/korpora/sich20/}.


\begin{table}[t!]
  \centering
  \setlength{\tabcolsep}{3pt}
    \begin{tabular}{lrrrrrrrrr}
    \toprule
    Class     & 1   & 2  &  3 & 4  & 5 &  6 & 7 & 8 & Sum \\  \midrule
    Frequency & 161 & 84 & 11 & 42 & 22 & 3 & 8 & 4 & 335 \\ 
    \bottomrule
    \end{tabular}
    \caption{Frequency distribution of \textit{sich} senses in manually annotated \textit{sich} dataset}
    \label{tab:class-freqs}
\end{table}

\section{Experimental Setup}

The specific word embedding model we employ is BERT
\cite{DBLP:journals/corr/abs-1810-04805}, a state-of-the-art
transformer. We use the `German BERT cased' model, which was trained
on a variety of German corpora, including Wikipedia, OpenLegalData,
and news articles \cite{ai19:_german_bert}. In comparison to the `BERT
multilingual' model which can also be used to model the semantics of
German text, the restriction of the training data to German leads in
particular to better tokenization. The model provides 768-dimensional
contextualized embeddings for all tokens presented to it as input.

We experiment with two conditions of presenting the \textit{sich}
instances in context to BERT. Recall that BERT learns contextualized
word embeddings -- that is, word embeddings that differ among
instances of the same word, reflecting the influence of context on
word meaning.  In the first condition, we present \textit{sich}
instances in their local \textit{phrasal} context, as approximated by
punctuation. That is, the context is formed by all words surrounding
\textit{sich} up to the closest commas, (semi)colons, or other phrasal
delimiters. The reason to use this oversimplification is that a proper
syntactic identification of the current phrase would have involved
full parsing of the sentences, which is still not possible at the
near-perfect accuracy we would require as a starting point for our
analysis.

In the second condition, we present them in their complete
\textit{sentential} context. To illustrate, the underlined part of the
following sentence makes up the phrasal context for the italicized
\textit{sich} (the English gloss is designed so as to match German
word order and punctuation):

\begin{quote}
Unsere Universität hat exzellent abgeschnitten und war auch nur indirekt -- aufgrund der landesweiten Unterauslastung -- lediglich in den 3 Bereichen Chemie, Physik und Slawistik, tangiert, \uline{die für \textit{sich} genommen allerdings ebenfalls exzellent dastehen}: [...] \\

`Our university has performed excellently and was only indirectly affected --
due to the countrywide underutilization -- only in the three areas of Chemistry, Physics and Slavic Studies, \uline{which, considered \textit{on their own}, however also appear excellent}: [...]'
\end{quote}
Our hypothesis is that the phrasal context provides a better basis for
distinguishing the senses of \textit{sich}, since its contents are of
higher average relevance. On the other hand, there is no guarantee
that our shallow definition of phrasal context captures all relevant
context cues. In the worst case, even the main verb may not be
present in the phrasal context, as the next example illustrates:
\begin{quote}
Abschließend lässt sich sagen, \textbf{dass \textit{sich} der Aufwand für diese Veranstaltung} (22 Stunden Zugfahrt an 2 Tagen für 2 Tage Seminar) insofern \uline{gelohnt} hat, [...] \\

`In sum, we conclude, \textbf{that \textit{\O} the effort for this
  event} (22 hours of train ride on 2 days for 2 days of workshop)
\uline{paid off} \textit{\O} insofar as [...]'
\end{quote}
This is why we also present \textit{sich} in the full sentence context.


\section{Exploratory Analysis} 
\label{sec:exploratory-analysis}

\begin{figure}[p]
    \centering 
     \includegraphics[width=0.95\textwidth]{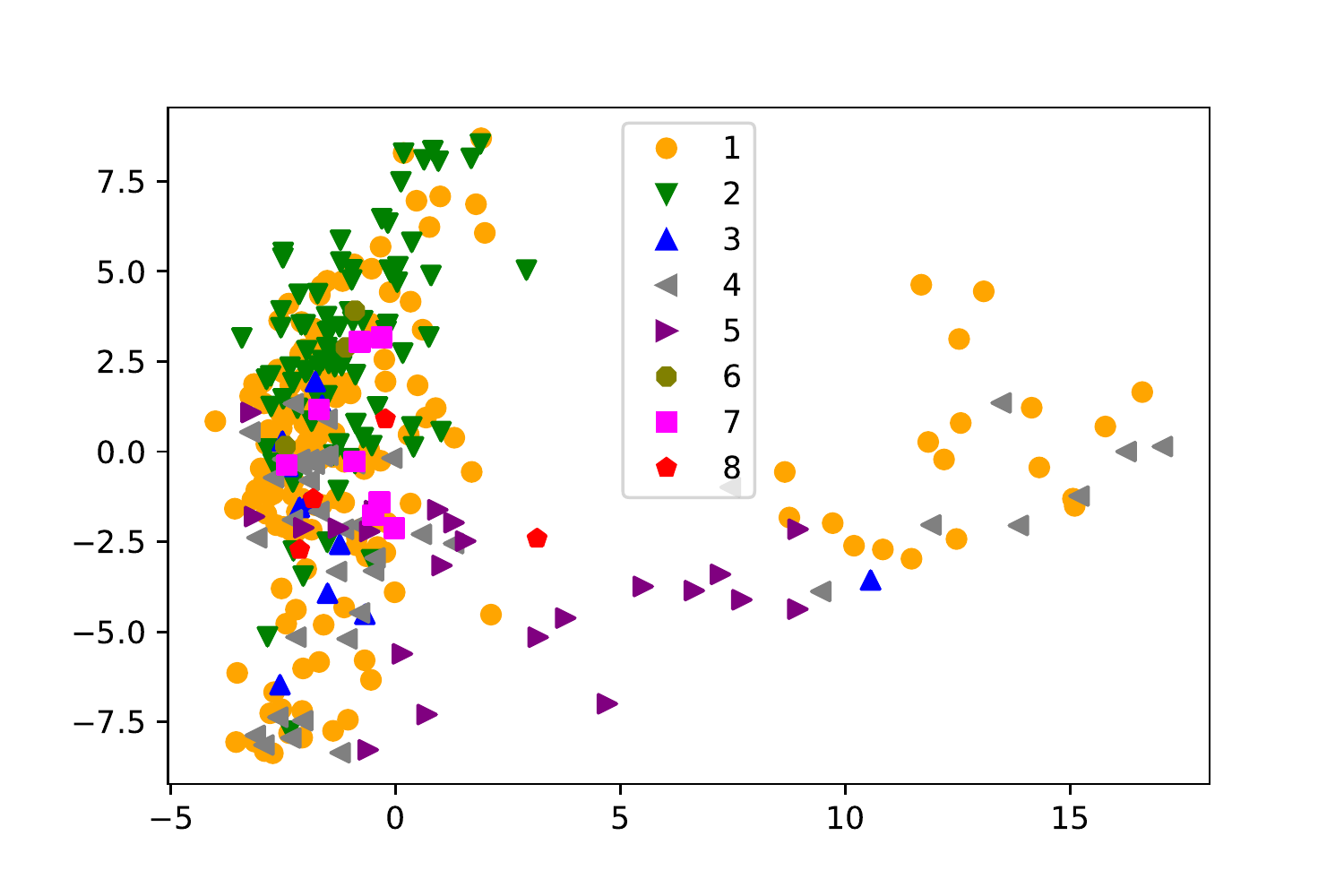}
       \includegraphics[width=0.95\textwidth]{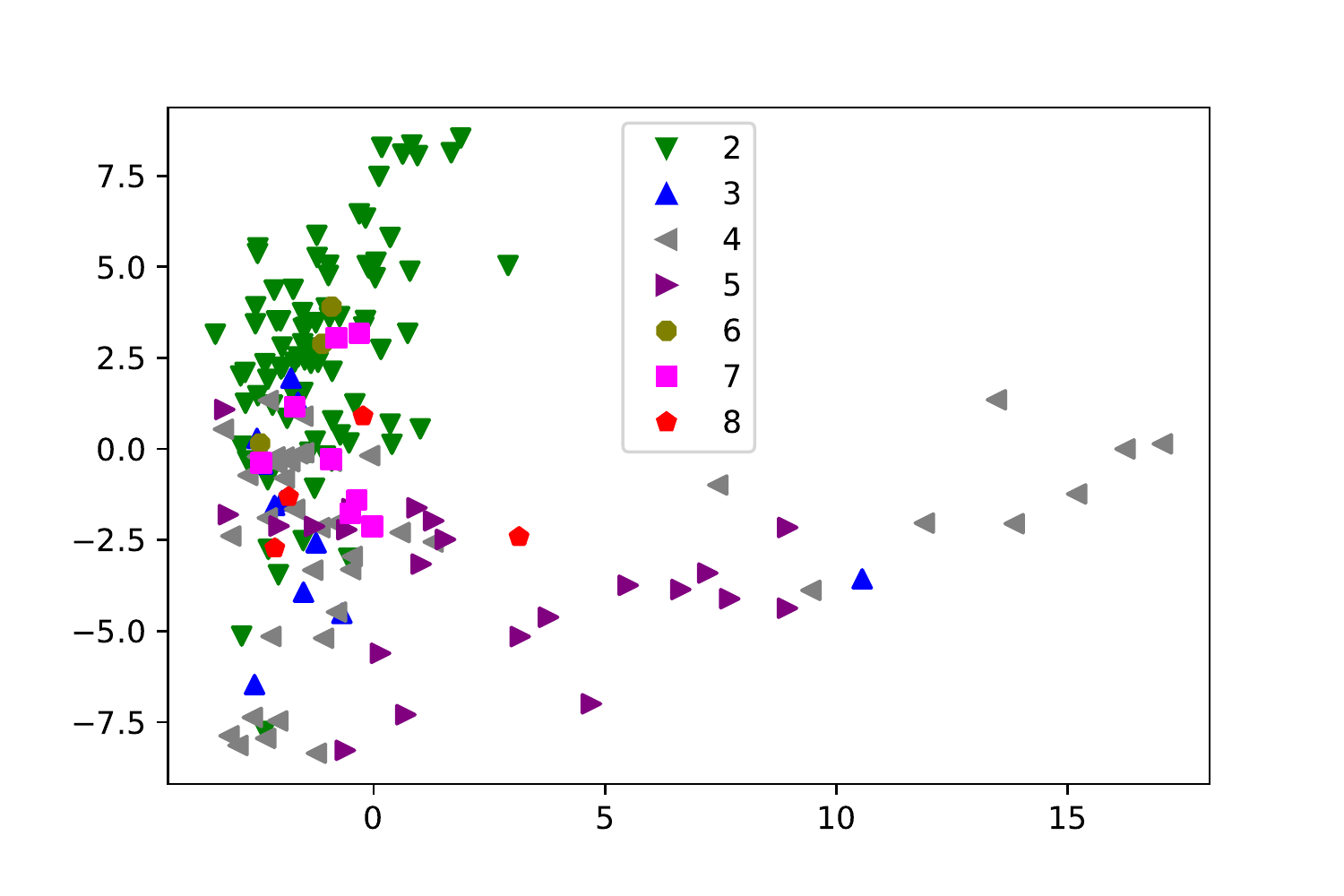}
       \caption{Distributional representations of \textit{sich}
         instances based on phrasal contexts. All classes (above),
         without inherent reflexives (below). Class labels according
         to Table~1.}
    \label{fig:fig1}
\end{figure}

As a first step, we perform an exploratory analysis in which we assess
to what extent we can recover the manually annotated senses in the
contextualized word embeddings produced by BERT when presented in
\textit{phrasal context}. We do so visually, by performing principal
components analysis (PCA), a dimensionality reduction method which
constructs a two-dimensional approximation of a higher-dimensional
space by capturing the directions of maximal variation
(i.e.,~differences among instances).  The result is a 2D
representation of our 335 \textit{sich} instances, as shown in
Figure~\ref{fig:fig1} (above: all classes, below: without Class~1).

%

In our estimation, the overall picture is promising. Even though the
classes are not completely separated, clear tendencies are
visible. Our observations are:

\begin{itemize}
\item Inherently reflexive verbs (Class~1) are interspersed through
  all event types and do not form a cluster of their own, as can be
  expected given their predictable nature. This motivates our showing
  a figure with Class~1 removed.
    
\item Typically other-directed reflexive events like `shooting
  oneself' and typically self-directed reflexive events like
  `defending oneself' or `combing' (Classes~4, 5) form neighboring
  categories in the bottom and right sectors.
    
\item The sectors at the bottom generally assemble agentive causative
  verb uses, whereas sectors in the top left corner assemble
  anticausative verb uses like `diminishing' or `revolving' (Class~2),
  which involve use of \textit{sich} in German.  Hence the gradient
  from top left to bottom right forms a path of growing agentivity,
  with traditional middle constructions (Classes~3, 6, 7) literally
  occupying the middle of the plot.
    
\item Some of the classes show a `core' surrounded by outlier
  clouds. For the change-of-posture verbs (Class~3), the outliers to
  the bottom and right are formed by the non-literal uses \textit{sich
    aus dem Verderben erheben} `to rise from doom' and \textit{sich
    auf die Rechtsgrundlage stützen} `to rest on the legal
  foundation').

\item The most inhomogeneous class is the class of self-directed verbs
  (Class~4), with one cluster in the mid-left sector and another on
  the right hand side. This can be explained in terms of the
  distinction between PP-\textit{sich} and DP-\textit{sich}
  \cite{gasthaas08}: The mid-left `core' of Class~4 consists of the DP
  cases, e.g. \textit{sich unterziehen} `to undergo'. In contrast, the
  outliers are made up of PP cases like \textit{bei sich tragen} `to
  carry'. The latter are clearly more causative, in line with the
  `causation' gradient described above.
\end{itemize}

\begin{figure}[tb]
    \centering
     \includegraphics[width=0.95\textwidth]{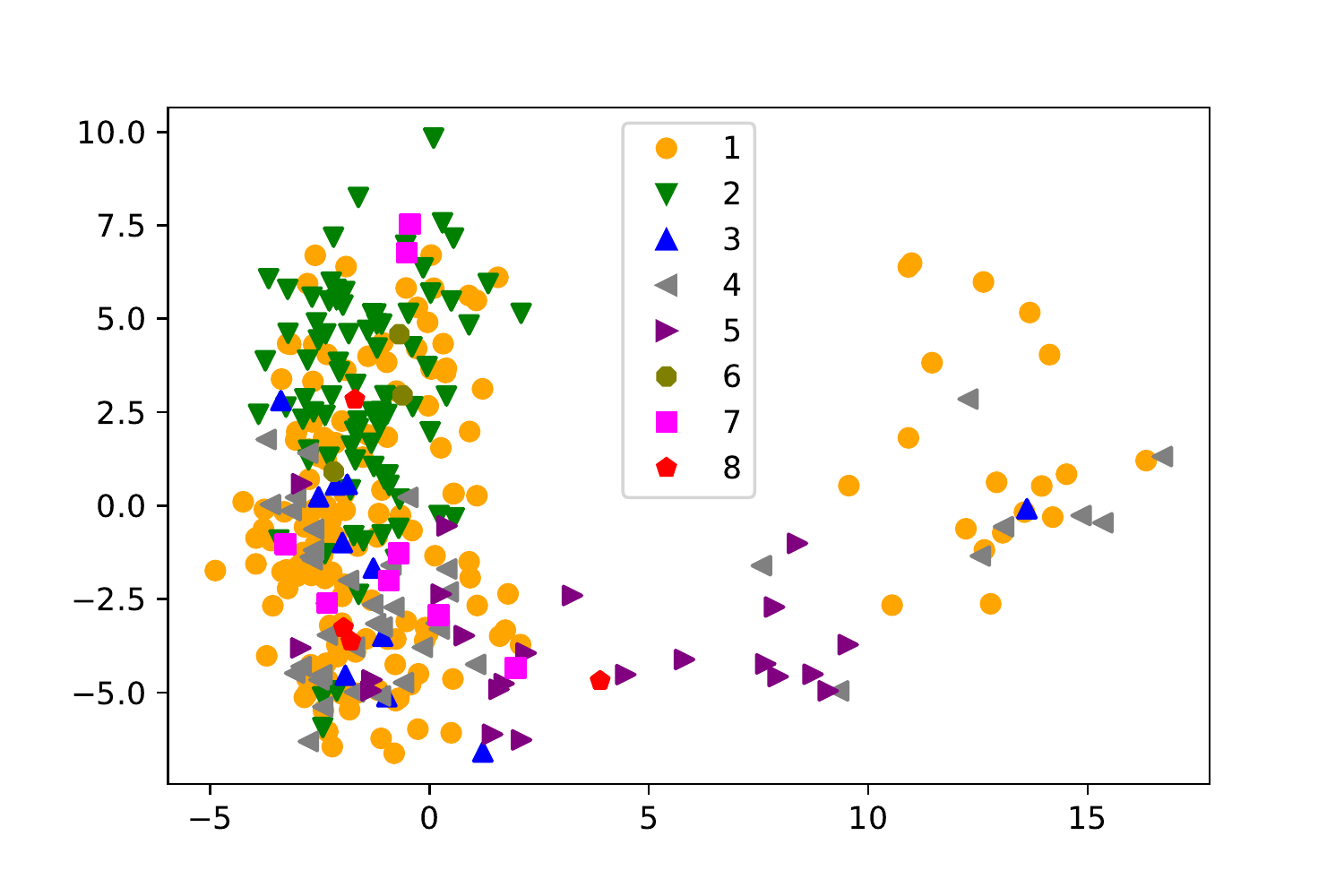}
    \caption{Distributional representations of \textit{sich} instances based on sentential contexts.}
    \label{fig:fig2}
\end{figure}

For comparison, Figure~\ref{fig:fig2} shows the instance embeddings
for \textit{sentential} contexts. The picture is overall similar to
the phrasal contexts. However, the clusters for the classes tend to be
even less tight than before, notably for Class 2 (for which we see
instances also at the bottom) and Class 7 (which also occurs at the
top). We interpret these observations as evidence for our hypotheses
stated above: the phrasal contexts -- which are on average 12 tokens
long -- are generally sufficient to disambiguate \textit{sich}, while
in the full sentential contexts -- which are on average 77 tokens long
-- the contribution of \textit{sich} is sometimes overwhelmed by the
topic of the complete sentence, as was observed in pre-transformer
distributional investigations. In the spirit of Occam's razor, we
focus on the phrasal context condition in the remainder of this
investigation.\footnote{We found a comparable, but slightly lower,
  performance for the sentential contexts in the classification
  experiments reported below. These experiments are part of the
  companion Jupyter notebook to this article.}



\section{Classification Experiments} 

The analysis in the previous section took into account only the 335
instances that we annotated manually. Naturally, it would be desirable
to scale up this analysis to large corpora and to automatically obtain
a large number of disambiguated \textit{sich} instances. In order to
do so, we trained a classifier which takes the contextualized
embeddings of \textit{sich} instances as input and returns one of the
eight senses as output. In essence, this classifier learns decision
boundaries between regions in embedding space that map onto different
classes.

To gauge the prospects for success in this procedure, we may inspect
Figure~\ref{fig:fig1}. Even though it is dangerous to draw strong
conclusions from dimensionality reduced visualizations (since there is
a loss of information compared to the original high-dimensional
vectors), it appears clear that Class 1 (inherent reflexives) follows
an essentially random distribution and will be hard to separate from
the other classes. For this reason, we carry out two experiments: one
where we consider all classes, and one where we leave Class 1 aside.
Finally, we report on an experiment that attempts to predict the
individual features of the classes.

\subsection{Experiment 1: Classification with All Classes}

For classification, we use a Support Vector Machine (SVM) with a
linear kernel, a standard choice of classification model.\footnote{We
  also experimented with fine-tuning the embeddings, but did not
  obtain competitive results, presumably due to the small size of the
  training set.} We perform 5-fold cross-validation, that is, we
divide the dataset into five partitions of 20\% each and run the model
five times, training on four partitions and evaluating on the fifth.

For evaluation, we apply the standard classification evaluation
measure, accuracy. As the percentage of correct predictions, accuracy
ranges between 0\% (all wrong) and 100\% (all correct). As a point of
comparison, we consider the \textit{most frequent class baseline}, the
accuracy achieved when always assigning the predominant
class. According to Table~\ref{tab:class-freqs}, Class~1 is the most
frequent class, with a relative frequency of 48.1\% -- that is, simply
assigning Class~1 to each datapoint would lead to an overall accuracy
of 48.1\%. Clearly, an informed model should outperform this baseline.

The SVM model, using the phrasal context, achieves an accuracy of
63.8\%. This result is some 15 points accuracy above the baseline, but
not even two out of three model's predictions are correct. This
indicates that the classification is relatively hard to make based on
the information present in the word embeddings. Table \ref{tab:exp1}
shows a simplified confusion matrix for Class~1 vs. all other classes,
where correct predictions are shown on the diagonal and incorrect
predictions off-diagonal. Indeed, this distinction is the main problem
of the classification. Most Class~1 instances are classified as such,
but more than one third of the instances of other classes are also
classified as Class~1. This is consistent with the classifier's
attempt to model the largest class (Class~1) as well as
possible. Unfortunately, this also means that the smaller classes are
not modeled appropriately.

\begin{table}[t!]
    \centering
    \begin{tabular}{llrr}
      \toprule
&       & \multicolumn{2}{c}{Predicted} \\ \cmidrule{3-4}
&                       & Class 1 & Other \\ \midrule
\multirow{2}{*}{Actual} & Class 1     &   129  &  32     \\
                        & Other      &    72   &  102     \\ 
\bottomrule
    \end{tabular}
    \caption{Experiment 1: Confusion matrix. Aggregated version: shows
      only Class~1 vs. all other classes. Overall accuracy of model:
      63.8\%.}
    \label{tab:exp1}
\end{table}

However, the blame should probably not fall entirely on the
classifier: As we saw in Section~\ref{sec:data-annotation}, the human
annotators also ran into problems to agree on some of the borderline
Class~1--Class~2 and Class~1--Class~3 cases, pointing towards the
inherent difficulty of these distinctions.

\subsection{Experiment 2: Classification without Inherent Reflexives}

Motivated by this finding, we tested in a second experiment how well
the other classes can be distinguished from one another. We adopted
the same setup as in Experiment~1 (SVMs with cross-validation), but
used only the 174 instances that were labeled as not Class~1 in the
gold standard.

This time, the classifier achieved an accuracy of 78.7\%, whereas the
most frequent class baseline is almost unchanged at 48.3\% (now the
most frequent class is Class 2). This is a clear improvement over the
accuracy shown in Experiment~1 -- the model outperforms the baseline
by 30 points accuracy. Clearly, the model is not perfect -- however,
its performance appears fair given the presence of ambiguous cases, as
discussed above.

The confusion matrix in Table \ref{tab:exp2} shows that the highest
numbers are indeed on the diagonal. In this setup, the hardest part of
the problem appears to be to distinguish Class~4 from Classes~2 and
3. This corresponds to our observations in
Section~\ref{sec:exploratory-analysis}, where we found Class~4 to be
represented in a relatively scattered manner due to its internal
heterogeneity (NP-sich vs. PP-sich, nonliteral cases).

\begin{table}[t!]
    \centering
    \begin{tabular}{llrrrrrrrr}
      \toprule
&       & \multicolumn{7}{c}{Predicted} \\ \cmidrule{3-9}
                        &        & Class 2 & Class 3 & Class 4 & Class 5 & Class 6 & Class 7 & Class 8 \\ \midrule
\multirow{7}{*}{Actual} & Class 2&77 & 1 & 5 & 1 & 0 & 0 & 0 \\
                        & Class 3& 3 & 2 & 5 & 1 & 0 & 0 & 0\\
                        & Class 4& 4 & 1 & 34&  3& 0 &  0&  0 \\
                        & Class 5& 1 & 0 &  5& 16& 0 &  0&  0 \\
                        & Class 6& 2 & 0 &  0&  0& 0 &  1& 0 \\
                        & Class 7& 0 & 0 &  0&  0& 0 &  8&  0 \\
                        & Class 8& 2 & 0 &  2&  0& 0 &  0&  0 \\
\bottomrule
    \end{tabular}
    \caption{Experiment 2: Confusion matrix for classification among
      all classes except Class~1. Full version. Overall model
      accuracy: 78.7\%)}
    \label{tab:exp2}
\end{table}

\subsection{Experiment 3: Prediction of Semantic Features}

A different approach towards distinguishing among the senses of
\textit{sich} is to consider these senses as bundles of features, as
defined in Table~\ref{tab:kemmer}. Concretely, this means that we can
predict the presence (or absence) of the five features from the
embeddings by phrasing them as binary classification tasks, again with
contextualized word embeddings as input. This approach enables us to
investigate whether any of these features are particularly
easy or difficult to predict.

We carried out this experiment for each of the features, using the
same experimental setup, model, and evaluation measure as in
Experiments~1 and 2. For each feature, we removed the instances for
classes which are neutral with regard to this feature ($\pm$ in
Table~\ref{tab:kemmer}) from consideration.

\begin{table}[t!]
  \centering
  \begin{tabular}{lrrrrr}
    \toprule
    Feature      & predictable & agentive & stressable& +\textit{lassen}& disposition \\
    \# instances & 335         & 159      & 331       & 335             & 86 \\
    Accuracy     & 80.0\%      & 88.6\%   & 95.4\%    & 99.4\%          & 96.5\% \\
    \bottomrule
  \end{tabular}
  \caption{Experiment 3: Prediction of individual semantic features}
  \label{tab:exp3}
\end{table}

The results are shown in Table~\ref{tab:exp3}, including the number of remaining
instances. Overall, the numbers look positive, with even the hardest
feature showing an accuracy of more than 80\% correct predictions.

The easiest feature to predict is `+\textit{lassen}', which is not
altogether surprising, given the obligatory presence of (an inflected
form of) \textit{lassen} in the context. In fact, the only error of
this classifier is an instance where \textit{lassen} was over ten
words away from \textit{sich}. The features `stressable' and
`disposition' are also relatively easy to predict ($>$ 90\%
accuracy). In the case of `disposition', this may be an effect of
correlation with `+\textit{lassen}', since, excluding the classes that
are neutral for this feature, the `disposition' instances are a strict
subset of the `+\textit{lassen}' instances. This interpretation is
bolstered by the observation that two of the three errors again
involve large distances between \textit{sich} and \textit{lassen}, as
above. It is interesting that `stressable' belongs to this category,
since stressability is a prosodic property that might not be reflected
directly in word embeddings, and arguably a property of the
construction rather than the individual instance.

The two features that are more difficult to predict are `agentive' and
`predictable'. Again, it is not surprising that `predictable' is a
hard feature, since this feature captures idiosyncratic, historically
fossilized properties of the predicate which, as we found over the
course of this article, are hard to capture for the embedding-based
methods we employed. There are also some borderline cases such as the
following:
\begin{quote}
  [...] wenn sie \textit{sich} redlich \uline{informiert} haben und vom geschichtlichen Hintergrund der Chilbi wissen [...] \\

  `If they have \uline{informed} \textit{themselves} honestly and know
  about the historical background of the Chilbi'
\end{quote}
We analysed this instance of \textit{sich informieren} `to inform
oneself' as an inherent reflexive (and thus `predictable') despite the
existence of the transitive \textit{jmd. informieren} `to inform
someone'. The reasons for our analysis are that \textit{sich} is
always unstressed in this collocation, and that \textit{$^*$er hat
  sich und andere informiert} `$^*$he informed himself and others' is
not possible, further evidence for the independence of the two
constructions. The classifier, however, did not reproduce our
analysis.
At any rate, these results tie in well with our observation in
Experiment~2 about the difficulty of distinguishing Class~4 from
Classes~2 and~3, which differ exactly with regard to these two
features.

Unfortunately, `recomposing' predictions for the individual features
into predictions about classes is not straightforward. The reason is
the partial neutrality of the classes with respect to the features,
which makes the mapping from features onto classes underspecified. For
example, an instance which is predictable, not agentive, not
stressable, without \textit{lassen}, and not dispositional, could
belong to either Class~1 or Class~2.

\section{Discussion and Conclusion} 

In this study, we have investigated the use of distributional meaning
representations to characterize the senses of a function word, the
German reflexive pronoun \textit{sich}. The main outcome of our study
is a positive one: the recent advances in distributional modeling of
lexical semantics, namely transformer-based contextualized embeddings,
have substantially increased the `resolution' of distributional
analysis: we can now characterize the meaning of function words not
only at the lemma level, but also at the level of individual
instances. In turn, this enables us to computationally model function
word polysemy and use the associated tools, such as visualization and
quantitative evaluation, to develop a better understanding of the
senses at hand.

An important limitation which we encountered in this study was that
one of the senses -- (meta-)Class 1, `inherent reflexives' -- turned
out to be rather difficult to distinguish from the other Classes, due
to the idiosyncratic behavior of its instances. This is an important
take-home message regarding the generalizability of our approach to
other function words or other phenomena in general: distributional
approaches, at the least in the incarnation we considered in this
study, are apt at capturing distinctions that can be grounded in
linguistic patterns, but they cannot account well for patterns that
are the result of historical fossilization.

This means that the classification setup that we used in the present
study, does not scale up directly to large corpora, as the results for
the other classes would be polluted by instances of Class 1, and vice
versa.  Note that this negative result hinges on the fact that we used
the standard formulation of classification, where we force the model
to assign a class to each and every instance. In view of the very
large number of attested \textit{sich} instances, which number 5.5
million in the SdeWAC corpus alone, this may not be the best
approach. A promising avenue for future work appears to be
experimenting with classifiers that only assign a class to instances
that they are very confident about. These `high-precision, low-recall'
classifiers would stand a better chance at identifying `prototypical'
instances of the various classes (maybe with the exception of Class~1)
and should still be able to collect substantial numbers for each
class. Evaluating such an approach would however require annotating
another sample of \textit{sich} instances, based on the confidence
estimates of the classifiers for the various classes.

Our present study can be compared and contrasted to another recent
study which investigated to what extent word embeddings encode world
knowledge attributes such as countries' areas, economic strengths, or
olympic gold medals \cite{gupta15:_distr}. The findings of that study
were remarkably similar to the present one in that the result was also
overall positive, but the difficulty of individual attributes was
directly related to the extent to which these attributes correlate
with salient patterns of linguistic usage in the underlying newswire
corpus -- high for area, low for olympic gold medals. Taken together,
these observations reaffirm the tight interactions between linguistic
and referential considerations in forming language, and the difficulty
of distinguishing between them in distributional analysis.

\bibliographystyle{splncs04}
\bibliography{sich}

\begin{thebibliography}{10}
\providecommand{\url}[1]{\texttt{#1}}
\providecommand{\urlprefix}{URL }
\providecommand{\doi}[1]{https://doi.org/#1}

\bibitem{bannard2003distributional}
Bannard, C., Baldwin, T.: Distributional models of preposition semantics. In:
  Proceedings of the ACL-SIGSEM Workshop on the Linguistic Dimensions of
  Prepositions and their Use in Computational Linguistics Formalisms and
  Applications. pp. 169--180. Toulouse, France (2003)

\bibitem{10.5555/2380816.2380822}
Baroni, M., Bernardi, R., Do, N.Q., Shan, C.c.: Entailment above the word level
  in distributional semantics. In: Proceedings of EACL. p. 23–32. Avignon,
  France (2012)

\bibitem{baroni-dinu-kruszewski:2014:P14-1}
Baroni, M., Dinu, G., Kruszewski, G.: Don't count, predict! a systematic
  comparison of context-counting vs. context-predicting semantic vectors. In:
  Proceedings of ACL. pp. 238--247. Baltimore, Maryland (2014)

\bibitem{bernardi-etal-2013-relatedness}
Bernardi, R., Dinu, G., Marelli, M., Baroni, M.: A relatedness benchmark to
  test the role of determiners in compositional distributional semantics. In:
  Proceedings of ACL. pp. 53--57. Sofia, Bulgaria (2013)

\bibitem{J12-3005}
Boleda, G., {Schulte im Walde}, S., Badia, T.: Modeling regular polysemy: A
  study on the semantic classification of {C}atalan adjectives. Computational
  Linguistics  \textbf{38}(3),  575--616 (2012)

\bibitem{cimiano2005learning}
Cimiano, P., Hotho, A., Staab, S.: {Learning Concept Hierarchies from Text
  Corpora using Formal Concept Analysis}. Journal of Artificial Intelligence
  Research  \textbf{24},  305--339 (2005)

\bibitem{ai19:_german_bert}
{Deepset.AI}: German {BERT} (2019), \url{https://deepset.ai/german-bert}

\bibitem{DBLP:journals/corr/abs-1810-04805}
Devlin, J., Chang, M., Lee, K., Toutanova, K.: {BERT:} pre-training of deep
  bidirectional transformers for language understanding. In: Proceedings of
  NAACL. pp. 4171--4186. Minneapolis (2019)

\bibitem{faas-13}
Faa{\ss}, G., Eckart, K.: {SdeWaC} -- a corpus of parsable sentences from the
  web. In: Language Processing and Knowledge in the Web, LNCS, vol.~8105, pp.
  61--68. Springer (2013)

\bibitem{Firth1957}
Firth, J.R.: {Papers in linguistics 1934-1951}. Oxford University Press (1957)

\bibitem{gasthaas08}
Gast, V., Haas, F.: On reciprocal and reflexive uses of anaphors in german and
  other european languages. In: König, E., Gast, V. (eds.) Reciprocals and
  reflexives: Theoretical and typological explorations, pp. 307--346. Mouton de
  Gruyter (2008)

\bibitem{gupta15:_distr}
Gupta, A., Boleda, G., Baroni, M., Pad\'o, S.: Distributional vectors encode
  referential attributes. In: Proceedings of EMNLP. Lisbon, Portugal (2015)

\bibitem{harris1954distributional}
Harris, Z.S.: Distributional structure. Word  \textbf{10}(2--3),  146--162
  (1954)

\bibitem{jawahar-etal-2019-bert}
Jawahar, G., Sagot, B., Seddah, D.: What does {BERT} learn about the structure
  of language? In: Proceedings of ACL. pp. 3651--3657. Florence, Italy (2019)

\bibitem{Kemmer-1993}
Kemmer, S.: The Middle Voice, Typological Studies in Language, vol.~23. John
  Benjamins, Amsterdam and Philadelphia (1991)

\bibitem{kappaInterpretation}
Landis, J.R., Koch, G.G.: The measurement of observer agreement for categorical
  data. Biometrics  \textbf{33}(1),  159--174 (1977),
  \url{http://www.ncbi.nlm.nih.gov/pubmed/843571}

\bibitem{levy2014neural}
Levy, O., Goldberg, Y.: Neural word embedding as implicit matrix factorization.
  In: Proceedings of NeurIPS. pp. 2177--2185. Montréal, QC (2014)

\bibitem{schneider-preps-acl18}
Schneider, N., Hwang, J.D., Srikumar, V., Prange, J., Blodgett, A., Moeller,
  S.R., Stern, A., Bitan, A., Abend, O.: Comprehensive supersense
  disambiguation of {E}nglish prepositions and possessives. In: Proceedings of
  ACL. pp. 185--196. Melbourne, Australia (2018)

\bibitem{Turney2010}
Turney, P.D., Pantel, P.: {From Frequency to Meaning: Vector Space Models of
  Semantics}. Journal of Artificial Intelligence Research  \textbf{37}(1),
  141--188 (2010)

\bibitem{vaswani2017attention}
Vaswani, A., Shazeer, N., Parmar, N., Uszkoreit, J., Jones, L., Gomez, A.N.,
  Kaiser, {\L}., Polosukhin, I.: Attention is all you need. In: Proceedings of
  NeurIPS. pp. 5998--6008. Long Beach, CA (2017)

\end{thebibliography}

\end{document}